\documentclass{tADR2e}

\newcommand{\bo}{\mathbf{o}}

\newcommand{\bs}{\mathbf{s}}
\newcommand{\ba}{\mathbf{a}}

\newcommand{\st}{\bs_t}
\newcommand{\stp}{\bs_{t+1}}
\newcommand{\at}{\ba_t}
\newcommand{\op}{\mathcal{O}}

\usepackage{url}
\usepackage{multibib}
\usepackage[normalem]{ulem}
\usepackage{subfiles}
\usepackage{comment}

\usepackage{tikz}
\usetikzlibrary{bayesnet}
\usetikzlibrary{calc}
\usepackage{algorithm}
\usepackage{color}
\usepackage{subfigure}
\usepackage{wrapfig}

\usepackage{graphicx}

\usepackage[noend]{algpseudocode}
\begin{document}

\jvol{00} \jnum{00} \jyear{2013} \jmonth{January}


\title{Integration of Imitation Learning using GAIL and Reinforcement Learning using Task-achievement Rewards via Probabilistic Graphical Model}

\author{Akira Kinose$^{a}$$^{\ast}$\thanks{$^\ast$Corresponding author. Email: latex.helpdesk@tandf.co.uk
\vspace{6pt}} and Tadahiro Taniguchi$^{b}$\\\vspace{6pt}  $^{a}${\em{Taylor \& Francis, 4 Park Square, Milton Park, Abingdon, UK}};
$^{b}${\em{Institut f\"{u}r Informatik, Albert-Ludwigs-Universit\"{a}t, Freiburg,Germany}}\\\vspace{6pt}\received{v1.0 released January 2013} }
\author{Akira Kinose and Tadahiro Taniguchi}
\maketitle

\begin{abstract}
The integration of reinforcement learning (RL) and imitation learning (IL) is an important problem that has been studied for a long time in the field of intelligent robotics. 
RL optimizes policies to maximize the cumulative reward, whereas IL attempts to extract general knowledge about the trajectories demonstrated by experts, i.e., demonstrators. Because each of them has its own drawbacks, methods combining them and compensating for each set of drawbacks have been explored thus far.
However, many of the methods are heuristic and do not have a solid theoretical basis.
In this paper, we present a new theory for integrating RL and IL by extending the probabilistic graphical model (PGM) framework for RL, {\it control as inference}. 
We develop a new PGM for RL with multiple types of rewards called probabilistic graphical model for Markov decision processes with multiple optimality emissions (pMDP-MO).
Furthermore, we demonstrate that the integrated learning method of RL and IL can be formulated as a probabilistic inference of policies on pMDP-MO by considering the discriminator in generative adversarial imitation learning (GAIL) as an additional optimality emission. 
We adapt the GAIL and task-achievement reward to our proposed framework, achieving significantly better performance than policies trained with baseline methods.

\medskip
\medskip
\end{abstract}

\section{Introduction}
The integration of reinforcement learning (RL) and imitation learning (IL) is an important problem that has been studied for a long time in the field of intelligent robotics and machine learning~\cite{lin1992self,lin1991programming,silver2016mastering,brys2015reinforcement,levine2013guided,zhu2018reinforcement,merel2017learning,hamahata2008effective,le2018hierarchical,nair2018overcoming,hester2018deep,andreas2017modular,sun2017deeply,rhinehart2018deep}.
RL and IL are both well-studied machine learning problems. The goal is to provide robots with capabilities to optimize policies automatically.
RL is conventionally formalized as a reward maximization problem. An RL agent explores the optimal policy by maximizing the cumulative rewards. However, it requires many trials in most of the robotic control problems. Because they have high dimensional states-and-actions spaces, exploration and optimization becomes a difficult problem.
In contrast, IL can acquire behavior by mimicking the behaviors of experts. However, its performance depends on the expert, and there is no guarantee of acquiring an optimal policy.
As a result, RL and IL have their own drawbacks.

The research for integrating RL and IL has been conducted to overcome problems and make use of the learner's trial-and-error experience and expert's demonstrations.
However, most variations are designed heuristically, i.e., few have a sophisticated theoretical basis~\cite{hamahata2008effective}.
This tends to lead to the heuristic parameter tuning and endless exploration of the variants of frameworks for integration.

The main goal of this paper is to present the integration of reinforcement and imitation learning methods into a single theoretical framework, i.e., probabilistic graphical model and its inference. 
Such integration allows the learning agents to use expert demonstrations and agent trial and error in a synergistic manner, enabling them to perform complicated tasks, such as robotic control, more swiftly.
For this purpose, we developed the probabilistic graphical model (PGM) framework with multiple optimality emissions as simultaneous observations. This theory is based on a framework that considers 
RL as probabilistic inference (a.k.a. control as inference) on the PGM, which has been gradually attracting attention~\cite{attias2003planning,todorov2007linearly,kappen2012optimal,toussaint2009robot,rawlik2013stochastic,ziebart2010modeling}.
Thus, it is recognized that RL, which optimizes a policy by maximizing the accumulating reward in a Markov decision process (MDP), can be re-formulated as a probabilistic inference in the PGM.
By interpreting learning control problems as probability theory, we can connect common probabilistic inferences to a wide range of problems, such as robotic control. This connection allows as to use a variety of approximate inference methods and probabilistic programming tools for reinforcement learning flexibly as well. 

We propose a framework integrating RL and IL by formulating them on the PGM framework by introducing multiple optimality emissions into the PGM for RL. We call the new model the new PGM probabilistic graphical model for Markov decision process with multiple optimality emission (pMDP-MO).
We demonstrate that the integrated learning method of RL and IL can be formulated as a probabilistic inference maximizing multiple optimality by adapting the optimality of RL and IL to PGM with multiple optimality emission.
Furthermore, we propose to use a generative adversarial imitation learning (GAIL) discriminator \cite{ho2016generative} 
to calculate the optimality distributions directly from the demonstrations. 
GAIL is a popular state-of-the-art IL method, which demonstrates good results in complex high-dimensional control tasks.
The proposed method learns the optimal policy using maximum entropy RL with task-achievement and imitation rewards, which are calculated by the GAIL discriminator. This model can be regarded as an extension of GAIL as well as an extension of PGM for RL. The method is referred to as GAIL using task-achievement reward (TRGAIL).

The main contributions of this study are as follows:
\begin{itemize}
  \item We developed PGM with multiple optimality emissions called pMDP-MO and formulated RL with multiple-types of rewards as probabilistic inference in pMDP-MO.
  \item We proposed TRGAIL, which integrates RL and IL by using the task-achievement reward and imitation reward calculated by the GAIL discriminator as multiple emission distribution on PGM, and demonstrated the effectiveness through experiments.
\end{itemize}
We experimented the proposed method on robot manipulation tasks in the physics simulator and found that our proposed method is better at sampling efficiency and learning performance than the conventional learning method.


\section{Background}
\subsection{Reinforcement Learning as Probabilistic Inference}
    Our proposal is based on the idea of the extension of PGM for RL as probabilistic inference (a.k.a. control as inference).
    Therefore, in this section, we briefly introduce RL as probabilistic inference~\cite{levine2018reinforcement}.
    
    A Markov decision process (MDP) is defined as $\mathcal{M} = (\mathcal{S}, \mathcal{A}, P, r, T)$. $\mathcal{S}$ represents the state space.
    $\mathcal{A}$ represents the action space.
    $P : \mathcal {S} \times \mathcal {A} \times \mathcal { S } \rightarrow \mathbb { R }$ denotes the transition probability distribution when taking action $a$ in state $s$ and going to $s'$.
    $r : \mathcal { S \times A} \rightarrow \mathbb { R }$ is the reward function after taking action $a$ in state $s$.
    $T$ is the time horizon of the task.
    We define $\pi$ as a stochastic policy $\pi : \mathcal { S } \times \mathcal { A } \rightarrow [ 0,1 ]$, and $\pi_E$ as an expert policy. 
    The expert demonstrations $\tau_E$ is a set of trajectories sampled by policy $\pi_E$.
    Trajectory $\tau$ consists of a sequence of state and action pairs.
    
     RL aims to optimize the parameters of the policy that maximizes the expected total reward based on the objective function  
    \begin{equation}
    \theta^\star = \arg\max_\theta \sum_{t=1}^T \mathbb{E}_{(\bs_t,\ba_t) \sim p(\bs_t,\ba_t|\theta)}[r(\bs_t,\ba_t)]. \label{eq:rl}
    \end{equation}
    Recently, significant progress has been made with regard to RL in continuous action decision problems, such as video games or robot control tasks, by integrating deep learning \cite{mnih2013playing,mnih2015human,silver2016mastering}.


\begin{figure}[t]
    \begin{center}
        \includegraphics[width=0.85\linewidth]{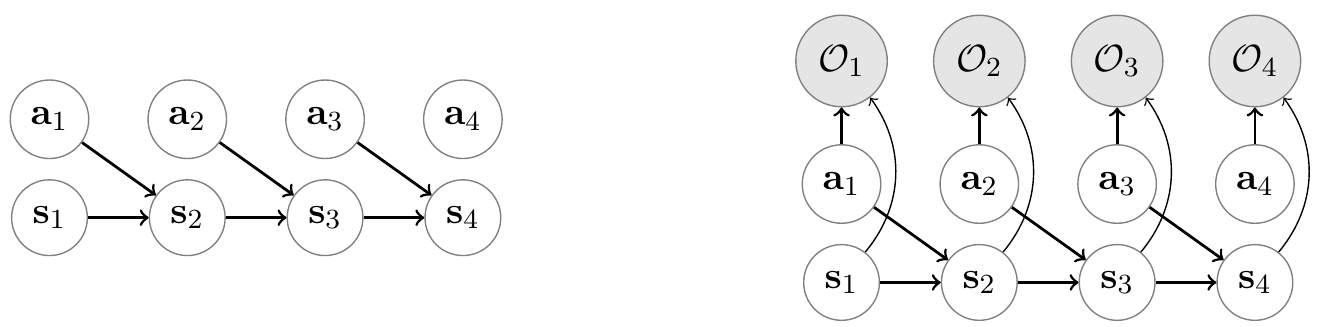}
      \caption{Left:  Probabilistic graphical model of MDP with states and actions as stochastic variables. Right: Probabilistic graphical model for reinforcement learning as probabilistic inference. The agent observes a single optimality $\op_t$ each timestep and infers the most probable action assuming that the optimality variables are taking true, i.e., $\op_t =1$. In this paper, we refer to the graphical model as a probabilistic graphical model from Markov decision process (pMDP).} 
      \label{fig:sub1}
  \end{center}
\end{figure}

As summarized by Levine in \cite{levine2018reinforcement}, the maximization of a reward in MDP can be re-formulated as the probabilistic inference problem in a PGM.
The study of RL as probabilistic inference (a.k.a. control as inference) is not new but rather has a decade of history \cite{attias2003planning,todorov2007linearly,kappen2012optimal,toussaint2009robot,rawlik2013stochastic,ziebart2010modeling}. Similar ideas have been proposed independently. 
Attias proposed an approach to the problem of planning under uncertainty using hidden Markov-style models \cite{attias2003planning}.
Todorov formulated a linear solvable Markov decision process that enables efficient approximation and determines the optimal policy efficiently \cite{todorov2007linearly}.
Kappen et al. reformulated Todorov's non-linear stochastic optimal control problem as a Kullback-Leibler (KL) minimization problem and efficiently calculated it by applying an approximate inference to the calculation of optimal control \cite{kappen2012optimal}.

In RL as probabilistic inference, the graphical model is constructed with optimality variables indicating whether or not a state-action pair is optimal.
Figure 1 illustrates the graphical model of MDP with an optimality variable.
The optimality variable in RL is denoted as
\begin{equation}
    p(\op_t = 1 | \bs_t, \ba_t) = \exp(r(\bs_t,\ba_t)).\label{eq:opdef}
\end{equation}
By introducing the optimality variable $\op_t$ as in Equation (\ref{eq:opdef}), RL can be formulated as probabilistic inference.
In RL as probabilistic inference, the policy is optimized by solving the probabilistic inference, such that the policy distribution is close to the optimal trajectory distribution conditioned to  $\op_t = 1$.

One way to optimize actions is as a particular type of structured variational inference.
We aim to approximate the posterior distribution over actions when we condition $\mathcal{O}_{t}=1$ for all $t \in\{1, \ldots, T\}$:
\begin {equation} 
    \begin{aligned} 
        p\left(\tau | \mathcal{O}_{1 : T}\right) \propto p\left(\tau, \mathcal{O}_{1 : T}\right) &=p\left(\mathbf{s}_{1}\right) \prod_{t=1} p\left(\mathcal{O}_{t}=1 | \mathbf{s}_{t}, \mathbf{a}_{t}\right) p\left(\mathbf{s}_{t+1} | \mathbf{s}_{t}, \mathbf{a}_{t}\right) \\ &=p\left(\mathbf{s}_{1}\right) \prod_{t=1}^{T} \exp \left(r\left(\mathbf{s}_{t}, \mathbf{a}_{t}\right)\right) p\left(\mathbf{s}_{t+1} | \mathbf{s}_{t}, \mathbf{a}_{t}\right)\\&=\left[p\left(\mathbf{s}_{1}\right) \prod_{t=1}^{T} p\left(\mathbf{s}_{t+1} | \mathbf{s}_{t}, \mathbf{a}_{t}\right)\right] \exp \left(\sum_{t=1}^{T} r\left(\mathbf{s}_{t}, \mathbf{a}_{t}\right)\right)
        \label{eq:dist1}
    \end{aligned}
\end {equation}

with another distribution:
\begin {equation} 
    \begin{aligned} 
        q(\tau)=q\left(\mathbf{s}_{1}\right) \prod_{t=1}^{T} q\left(\mathbf{s}_{t+1} | \mathbf{s}_{t}, \mathbf{a}_{t}\right) q\left(\mathbf{a}_{t} | \mathbf{s}_{t}\right).
    \label{eq:dist2}
    \end{aligned}
\end {equation}
By optimizing the variational lower bound, we can approximate the inference of an optimal trajectory distribution.
Using Equation \ref{eq:dist1}, \ref{eq:dist2}, and Jensen's inequality, the variational lower bound of the log-likelihood is given by
\begin {equation} 
    \begin{aligned} 
        \log p\left(\mathcal{O}_{1 : T}\right) &=\log \iint p\left(\mathcal{O}_{1 : T}, \mathbf{s}_{1 : T}, \mathbf{a}_{1 : T}\right) d \mathbf{s}_{1 : T} d \mathbf{a}_{1 : T} \\ 
        &=\log \iint p\left(\mathcal{O}_{1 : T}, \mathbf{s}_{1 : T}, \mathbf{a}_{1 : T}\right) \frac{q\left(\mathbf{s}_{1 : T}, \mathbf{a}_{1 : T}\right)}{q\left(\mathbf{s}_{1 : T}, \mathbf{a}_{1 : T}\right)} d \mathbf{s}_{1 : T} d \mathbf{a}_{1 : T} \\
        &=\log \mathbb{E}_{\left(\mathbf{s}_{1 : T}, \mathbf{a}_{1 : T}\right) \sim q\left(\mathbf{s}_{1 : T}, \mathbf{a}_{1 : T}\right)}\left[\frac{p\left(\mathcal{O}_{1 : T}, \mathbf{s}_{1 : T}, \mathbf{a}_{1 : T}\right)}{q\left(\mathbf{s}_{1 : T}, \mathbf{a}_{1 : T}\right)}\right] \\
        &\geq \mathbb{E}_{\left(\mathbf{s}_{1 : T}, \mathbf{a}_{1 : T}\right) \sim q\left(\mathbf{s}_{1 : T}, \mathbf{a}_{1 : T}\right)}\left[\log p\left(\mathcal{O}_{1 : T}, \mathbf{s}_{1 : T}, \mathbf{a}_{1 : T}\right)-\log q\left(\mathbf{s}_{1 : T}, \mathbf{a}_{1 : T}\right)\right] \\
        &=\mathbb{E}_{\left(\mathbf{s}_{1 : T}, \mathbf{a}_{1 : T}\right) \sim q\left(\mathbf{s}_{1 : T}, \mathbf{a}_{1 : T}\right)}\left[\sum_{t=1}^{T} r\left(\mathbf{s}_{t}, \mathbf{a}_{t}\right)-\log q\left(\mathbf{a}_{t} | \mathbf{s}_{t}\right)\right].
    \label{eq:dist3}
    \end{aligned}
\end {equation}
From the above, in the context of RL as inference, reward maximization by RL in MDP is equivalent to the objective function of maximum entropy RL~\cite{levine2018reinforcement}.

In recent studies, maximum a posteriori policy optimization, which is one of the RL methods serving as a probabilistic inference, shows good results by applying the EM algorithm to the problem of RL \cite{abdolmaleki2018maximum}.
Fu et al. formulated a probabilistic inference that maximizes the probability of the occurrence of events in the framework of control as inference and proposed a generalization of inverse reinforcement learning (IRL) methods to cases in which full demonstrations are not needed, such as when only samples of desired goal states are available \cite{fu2018variational}.

\subsection{Generative Adversarial Imitation Learning}
    We consider integrating GAIL into RL via the framework of PGM for RL.
    Therefore, in this section, we briefly summarize GAIL and IL.
    The goal of IL is to acquire the behavior that mimics expert behavior. 
    IL can be learned without reward signal $r$ when compared to RL, but it is necessary to prepare expert demonstrations in advance.
    IL can be classified into two approaches:
    \begin{enumerate}
        \item Behavior cloning (BC) \cite{bain1999framework,pomerleau1991efficient}, which learns policy over state-action pairs in supervised learning on expert demonstration $\tau_E$.
        BC has been successfully applied to autonomous driving \cite{bojarski2016end} and locomotions \cite{nakanishi2004learning,kalakrishnan2009learning}.
        The BC approach is difficult to use in the real world because of the compounding error caused by a covariate shift.
        \item Inverse reinforcement learning (IRL) \cite{ziebart2008maximum,wulfmeier2015maximum,finn2016guided,ho2016generative,fu2017learning} recovers the reward function under the assumption that the expert policy is optimal and learns the policy on the recovered reward function.
        However, inverse reinforcement learning is too expensive to perform because it requires solving an RL in its learning process loop.
    \end{enumerate}

    Recently, Ho and Ermon developed GAIL \cite{ho2016generative}, which is the IL method inspired by generative adversarial networks (GAN) \cite{goodfellow2014generative}.
    GAIL is able to imitate the policy for complex high-dimensional control tasks.
    In the GAIL framework, the agent imitates the behavior of the expert policy by matching the generated state-action distribution with the distribution of experts.
    
    GAIL's {\it generator} tries to make the {\it discriminator} recognize that the state-action pairs generated by the policy are generated from an expert.
    GAIL's discriminator distinguishes state-action pairs from those generated by the generator and expert.
    As learning progresses, the discriminator guides the policy to match the state-action pairs generated by the generator with the expert state-action pairs.
    When the Jensen-Shannon divergence between agent policy distributions and expert policy distribution is minimized, the agent policy is optimal under the condition that the expert policy is the optimal policy.
    
    GAIL's objective function is denoted as
    \begin{eqnarray}
    	\label{eq:GAILO}
        \max_{\theta} \min_{\omega} \mathbb{E}_{\pi_\theta}&[\log(D_\omega(s,a))]+\mathbb{E}_{\pi_{E}}& [\log(1-D_\omega(s,a))]
    \end{eqnarray} 
    where $\pi_\theta$ is the agent policy that is the role of the generator, $\pi_E$ is the expert policy, and $D_\omega$ is a discriminator that tries to distinguish state-action pairs generated from $\pi_\theta$ and $\pi_E$.
    In other words,  $D_\omega$  outputs the probability that the state-action pair is optimal under the assumption that the demonstrator's behavior is optimal.
    Parameters $\theta$ and $\omega$ are the parameters of the generator and discriminator, respectively, which are represented as the deep neural network.
    The generator is trained by a policy gradient method, such as trust region policy optimization \cite{schulman2015trust} and proximal policy optimization (PPO) \cite{schulman2017proximal}.
    The discriminator is optimized using ADAM \cite{kingma2014adam}. Some variants of GAIL have been proposed recently as well~\cite{baram2017end,li2017infogail,hausman2017multi,lin2018acgail}.

\subsection{Integration of Reinforcement Learning and Imitation Learning via PGM}
RL and IL are both well-established approaches; however, each has its own drawbacks.
On the one hand, RL has the following drawbacks.
First, it is difficult to manually design the appropriate reward function in complicated or high-dimensional tasks, such as robotic control.
Second, the computational cost of learning is prohibitively expensive because of the exploration of the policy space for the reward.

On the other hand, IL has the following drawbacks.
First, its performance depends on the expert, with no guarantee of acquiring the optimal policy. 
Second, if the number of expert demonstrations is small, it is difficult for agents to learn adequately.
Third,  Most IL methods cannot effectively use environmental feedbacks.

    Several conventional works have attempted to improve performance by combining RL with the learning from demonstrations~\cite{lin1992self,silver2016mastering,brys2015reinforcement,levine2013guided}.
    By adapting RL to the policy learned from demonstrations, it is possible to avoid searching for unnecessary action space. The combination of RL and learning from demonstration can evaluate whether the task is performed well, such that the learned policy is improved.     
    
    In an early work, Lin used a successful demonstration to improve RL more efficiently in a 2D-dynamics game \cite{lin1992self}.
    The use of demonstrations becomes more effective as the task complexity increases \cite{lin1991programming}.
    The most famous approach of imitation and reinforcement learning is AlphaGo, which was used to learn the game Go and was proposed by Silver et al \cite{silver2016mastering}.
    
    As a different approach to the combination of RL and IL, Brys used demonstrations as a prior knowledge for the formation of reward functions \cite{brys2015reinforcement}.
    This approach of using expert demonstrations to form a reward function is similar to that of inverse reinforcement learning.
    Levine and Koltun generated guide samples from human demonstrations and used them to explore high reward areas of the policy space \cite{levine2013guided}.
    In recent years, many approaches have been proposed to improve performance by combining deep reinforcement learning and IL\cite{zhu2018reinforcement,merel2017learning,hamahata2008effective,le2018hierarchical,nair2018overcoming,hester2018deep,andreas2017modular,sun2017deeply,rhinehart2018deep}
    
    However, most of the integration was performed in a heuristic manner and was not formulated on a single probabilistic generative model.
     The main goal of this paper is to present a theoretical learning framework that can be used in complicated tasks, such as robotic control, by integrating these methods through pMDP-MO and making use of the GAIL discriminator as an optimality distribution.

\section{Probabilistic Graphical Model for Markov Decision Process with Multiple Optimality Emissions (pMDP-MO)}

\begin{figure}[]
    \centering
        \includegraphics[width=0.4\linewidth]{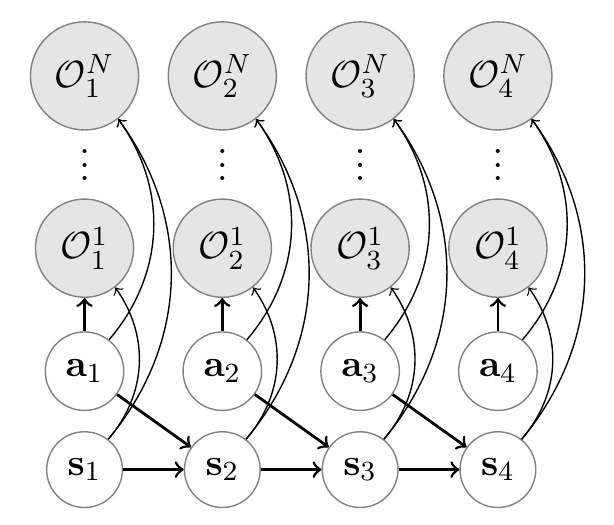}
\caption{Graphical model framework for reinforcement learning as probabilistic inference with multiple optimal emissions. In this framework, the agent observes multiple optimalities $\op_t^1, ..., \op_t^N$ and infers the optimal behavior.}
      \label{fig:sub2}
\end{figure}

In this section, we develop the new PGM framework for representing MDP with simultaneous observation of multiple optimality emissions, called pMDP-MO.
This graphical model is obtained by extending the graphical model framework for control as inference, as shown in Figure \ref{fig:sub1}.
We assume a Markov decision process model in which multiple optimalities for the state-action pair at the time $t$ are observed simultaneously.

Assuming that the respective optimality variables are independent of each other, the probability distribution that multiple optimality emissions are observed is described as follows:
\begin{align}
    p(\op_{t}^{1},\op_{t}^{2}, \ldots  ,\op_{t}^{N}|\bs_t,\ba_t) = \prod_{n=1}^N p(\op^{n}|\bs_t,\ba_t) 
\end{align}
where $N$ is the number of optimality types observed simultaneously and $\op_{t}^{n}$ is the $n$-th optimality.
Figure \ref{fig:sub2} depicts a graphical model in this setting in which multiple types of optimality are observed simultaneously.

Based on PGM, we can derive the optimal trajectory probabilistic distribution $p(\tau | \op_{1:T}^1, \ldots  ,\op_{1:T}^N) $ by considering that the state-action pairs are optimal $\op_t^1 = 1,  \ldots   , \op_t^N = 1$ at time to horizon $T$.
\begin{align}
    p(\tau | \op^1_{1:T},...,\op^N_{1:T}) \propto p(\tau, \bo^1_{1:T},...,\bo^N_{1:T})= p(\bs_1)\prod_{t=1}^T \left[ p(\bs_{t+1}|\bs_t,\ba_t) \prod_{n=1}^N p(\op^{n}|\bs_t,\ba_t)\right]  \label{eq:3.1.1}
\end{align}

Using Equation \ref{eq:dist1}, \ref{eq:3.1.1} and Jensen's inequality, the variational lower bound of log-likelihood is given by
\begin {equation} 
    \begin{aligned} 
        \log p\left(\mathcal{O}^1_{1 : T}, \ldots ,\mathcal{O}^N_{1 : T}\right) &=\log \iint p\left(\mathcal{O}^1_{1 : T},\ldots,\mathcal{O}^N_{1 : T}, \mathbf{s}_{1 : T}, \mathbf{a}_{1 : T}\right) d \mathbf{s}_{1 : T} d \mathbf{a}_{1 : T} \\ 
        &=\log \iint p\left(\mathcal{O}^1_{1 : T}, \ldots  ,\mathcal{O}^N_{1 : T}, \mathbf{s}_{1 : T}, \mathbf{a}_{1 : T}\right) \frac{q\left(\mathbf{s}_{1 : T}, \mathbf{a}_{1 : T}\right)}{q\left(\mathbf{s}_{1 : T}, \mathbf{a}_{1 : T}\right)} d \mathbf{s}_{1 : T} d \mathbf{a}_{1 : T} \\
        &=\log \mathbb{E}_{\left(\mathbf{s}_{1 : T}, \mathbf{a}_{1 : T}\right) \sim q\left(\mathbf{s}_{1 : T}, \mathbf{a}_{1 : T}\right)}\left[\frac{p\left(\mathcal{O}^1_{1 : T}, \ldots  ,\mathcal{O}^N_{1 : T}, \mathbf{s}_{1 : T}, \mathbf{a}_{1 : T}\right)}{q\left(\mathbf{s}_{1 : T}, \mathbf{a}_{1 : T}\right)}\right] \\
        &\geq \mathbb{E}_{\left(\mathbf{s}_{1 : T}, \mathbf{a}_{1 : T}\right) \sim q\left(\mathbf{s}_{1 : T}, \mathbf{a}_{1 : T}\right)}\left[\log p\left(\mathcal{O}^1_{1 : T}, \ldots ,\mathcal{O}^N_{1 : T}, \mathbf{s}_{1 : T}, \mathbf{a}_{1 : T}\right)-\log q\left(\mathbf{s}_{1 : T}, \mathbf{a}_{1 : T}\right)\right] \\
       &= \mathbb{E}_{\left(\mathbf{s}_{1 : T}, \mathbf{a}_{1 : T}\right) \sim q\left(\mathbf{s}_{1 : T}, \mathbf{a}_{1 : T}\right)}\sum_{t=1}^{T}\left[ \sum_{n=1}^{N} \log p(\op_t^n| \bs_t, \ba_t) -\log q\left(\mathbf{s}_{t}| \mathbf{a}_{t}\right)\right] \\
    \label{eq:3.1.2}
    \end{aligned}
\end {equation}
This is derived by using the same procedure as Equation \ref{eq:dist3}.
Thus, in the PGM framework for simultaneously observing multiple optimality emissions, the probability that two optimalities are observed is denoted as Equation \ref{eq:3.1.2}.

If the distribution of optimality follows Equation \ref{eq:opdef}, i.e., $p(\op_t^n| \bs_t, \ba_t) = \exp(r^n(\bs_t,\ba_t))$, the variational lower bound of log-likelihood is given by
\begin {equation} 
    \begin{aligned} 
        \log p\left(\mathcal{O}^1_{1 : T}, \ldots ,\mathcal{O}^N_{1 : T}\right)
        &\geq \mathbb{E}_{\left(\mathbf{s}_{1 : T}, \mathbf{a}_{1 : T}\right) \sim q\left(\mathbf{s}_{1 : T}, \mathbf{a}_{1 : T}\right)}\sum_{t=1}^{T}\left[ \sum_{n=1}^{N}r^n\left(\mathbf{s}_{t}, \mathbf{a}_{t}\right)
       -\log q\left(\mathbf{s}_{t}| \mathbf{a}_{t}\right)\right]. \\
     \label{eq:3.1.3}
    \end{aligned} 
\end {equation} 

Equation \ref{eq:3.1.3} shows that assuming multiple types of optimalities is the same as providing several types of sub-rewards in an additive manner.  

\section{GAIL using Task-achievement Reward (TRGAIL)}
In this section, we show that the integrated learning of RL and IL could be formulated as the maximum entropy reinforcement learning for optimality for the reward in RL, i.e., task-achievement reward, and optimality for IL, representing whether the action is similar to that of the demonstrator. The probability for the latter is calculated by the GAIL discriminator.

We define two types of optimality, optimality for RL $\op_t^R$  and  optimality for IL $\op_t^I$ .
We define the distribution over $\op_t^R$ as
    \begin{equation}
    p(\op_t^R = 1 | \bs_t, \ba_t) = \exp(r(\bs_t,\ba_t)).\label{eq:opgdef}
    \end{equation}
By using this formulation, as with general RL, the agent learns a policy to maximize the accumulated expected designed rewards  $r$ in the PGM framework (see section 2.1). This definition of optimality represents the target of RL, i.e., maximizing cumulative rewards. 

In contrast, $\op_t^I$ represents the goal of IL. The goal of IL is to simulate expert behavior, which is assumed to be optimal.
We adopt GAIL discriminator $D_\omega(\bs_t,\ba_t)$, which indicates the probability if the state and action pair is generated from experts, i.e., the optimal controller, to calculate  $p(\op_t^I = 1 | \bs_t, \ba_t)$ ie. this can be interpreted as the optimality for imitation.
    \begin{equation}
    p(\op_t^I = 1 | \bs_t, \ba_t) = \exp(\log(D_\omega(\bs_t,\ba_t))) = D_\omega(\bs_t,\ba_t).\label{eq:opddef}
    \end{equation}
Increasing the probability $p(\op_t^I = 1 | \bs_t, \ba_t)$ implies that the agent policy is updated to be close to the expert policy.
The discriminator is parameterized with the parameter $\omega$ and learned alternately with optimality maximization by the generator in the same way as GAIL.

From Equation (\ref{eq:3.1.1},\ref{eq:3.1.2},\ref{eq:opgdef},\ref{eq:opddef}), the variational lower bound of log-likelihood of optimality is derived as
\begin{align*}
    \log p(\op^R_{1:T},\op^I_{1:T}) \geq  \mathbb{E}_{(\bs_{1:T},\ba_{1:T})\sim q(\bs_{1:T}, \ba_{1:T})} \left[ \sum_{t=1}^T r(\bs_t,\ba_t) + \log(D_\omega(\bs_t,\ba_t)) - \log q(\ba_t|\bs_t) \right].
\end{align*}
From the above, the integrated learning of RL and IL is formulated as the maximum entropy reinforcement learning in which the reward function is defined as $r(\bs_t,\ba_t) + \log(D_\omega(\bs_t,\ba_t))$.

In this study, we use PPO \cite{schulman2017proximal}, which is an RL method of policy gradient, to maximize the reward function $r(\bs_t,\ba_t) + \log(D_\omega(\bs_t,\ba_t))$. Note that other RL algorithms can be used in TRGAIL. 
By adding the policy entropy maximization term to the objective function of the actor-part of the policy gradient, we regard it as maximum entropy reinforcement learning.
Therefore, the generator maximizes reward $r(\bs_t,\ba_t) + \log(D_\omega(\bs_t,\ba_t))$ and the entropy of the policy.
The discriminator is trained to discriminate between the trajectory of the expert demonstration and the trajectory generated from the generator.

In this study, we assume the task-achievement reward as a reward for RL. It is difficult to solve RL with a simple task-achievement reward function. Therefore, in many studies of RL, designing the rewards is a problem to be solved, and researchers manually designed reward functions to facilitate RL.  
TRGAIL aims to solve the problem by making use of the trajectories given by demonstrators, i.e., by integrating imitation learning.

The task-achievement reward is the binary reward that indicates whether the current state-action pair achieved the task.
Based on the task-achievement reward and imitation reward trained by GAIL, we refer to this method of learning as the generative adversarial imitation learning using task-achievement reward (TRGAIL).

\begin{figure}[!t]
    \begin{center}
      \includegraphics[width=13cm]{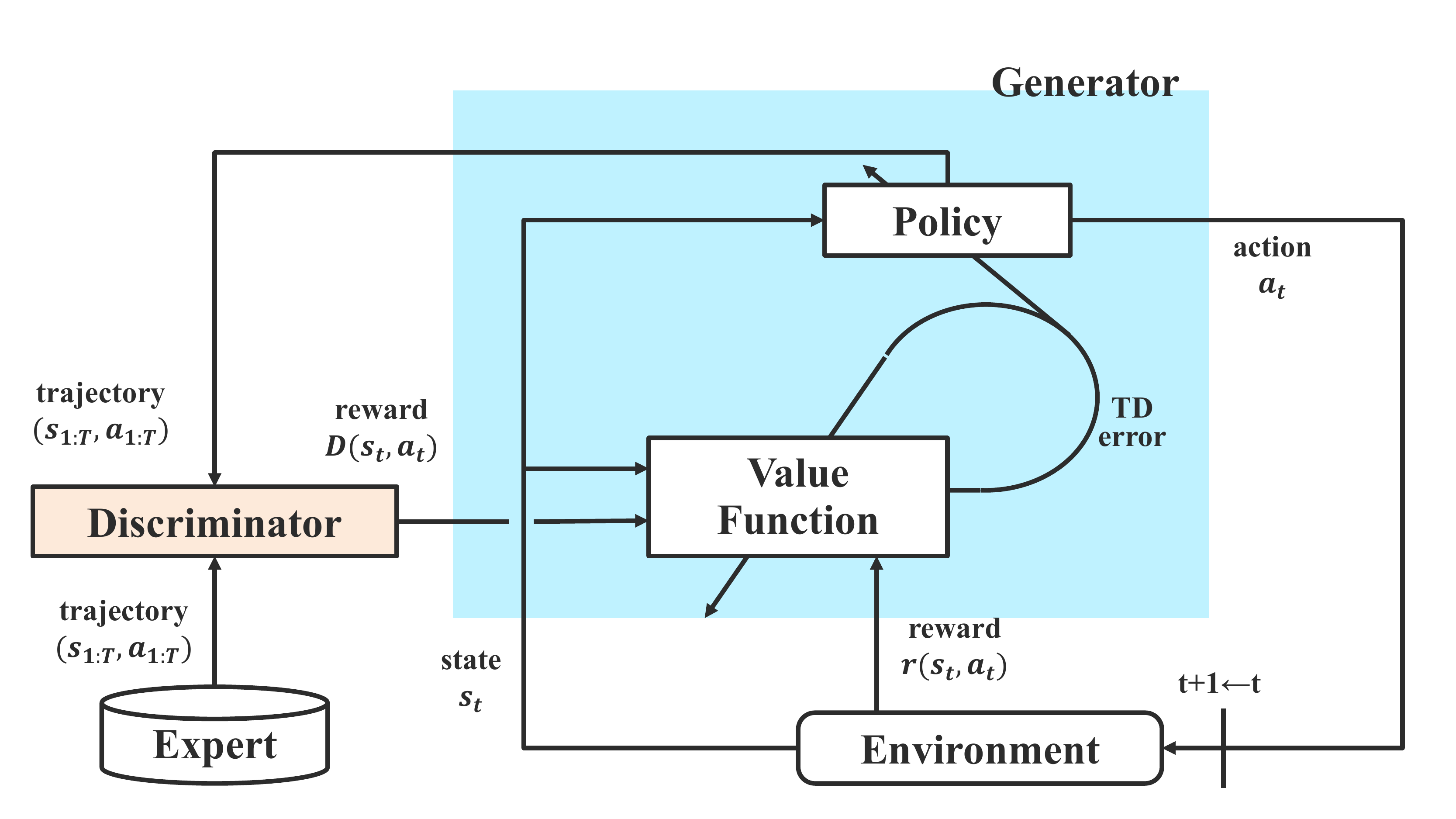}
      \label{proposed_model_pic}
      \caption{\textbf{Model structure of the proposed method.} The agent receives rewards from both the discriminator and the environment, and trains the value function, policy, and discriminator.}
    \end{center}
\end{figure}

Figure \ref{proposed_model_pic} depicts the model structure of TRGAIL. 
Algorithm 1 shows the algorithm flow of the TRGAIL.

\renewcommand{\algorithmicrequire}{\textbf{Input:}}
\renewcommand{\algorithmicensure}{\textbf{Output:}}

\begin{algorithm}[tb]
    \caption{GAIL using task-achievement reward}
    \label{alg:gail}
    \begin{algorithmic}
       \State {\bfseries Input:} $\tau_E \sim \pi_E$, \Comment{Sample expert's trajectories} 
       \State {\bfseries Input:} $\theta, \omega $\Comment{Initialize network parameters}
       \For{each iteration $i$}
            \For{each environment step $t$}
                \State $\at \sim \pi_\theta(\at|\st)$ \Comment{Sample action from the agent policy}
                \State $\stp \sim p(\stp| \st, \at)$ \Comment{Sample transition from the environment}
    	        \State $\mathcal{D} \leftarrow \mathcal{D} \cup \left\{(\st, \at, r(\st, \at), \stp)\right\}$ \Comment{Store the transition in the replay buffer}
    	    \EndFor
        \For{each discriminator gradient step}
            \State Sample trajectories $\tau_i \sim \mathcal{D}$
            \State $w_i \leftarrow w_i - \lambda_D \left( \hat{\mathbb{E}}_{\tau_i}[\nabla_w  \log(D_w(s,a))] + \hat{\mathbb{E}}_{\tau_E}[\nabla_w  \log(1-D_w(s,a))]\right)$ \\
            \Comment{Update the discriminator parameters}
        \EndFor
        \For{each generator gradient step}
    	    \State $\theta_i \leftarrow \theta_i + \lambda_\pi \mathbb{E}_{\pi} \left[ \sum_{t=1}^T r(\bs_t,\ba_t) + \log(D_\omega(\bs_t,\ba_t)) - \log \pi_\theta(\ba_t|\bs_t) \right]$\\
    	    \Comment{Update the policy parameters}
        \EndFor
    \EndFor
    \end{algorithmic}
\end{algorithm}

\section{Experiment}
To evaluate our algorithm, we used three physics-based control tasks--Pusher, Striker, and Thrower--which are simulated using the MuJoCo physics simulator \cite{todorov2012mujoco}.
Then, using a number of trajectories generated by the expert policy, we trained our algorithm.

\subsection{Conditions}
\begin{figure*}[!t]
    \subfigure[Pusher Task]{
        \includegraphics[clip,width=0.33\linewidth]{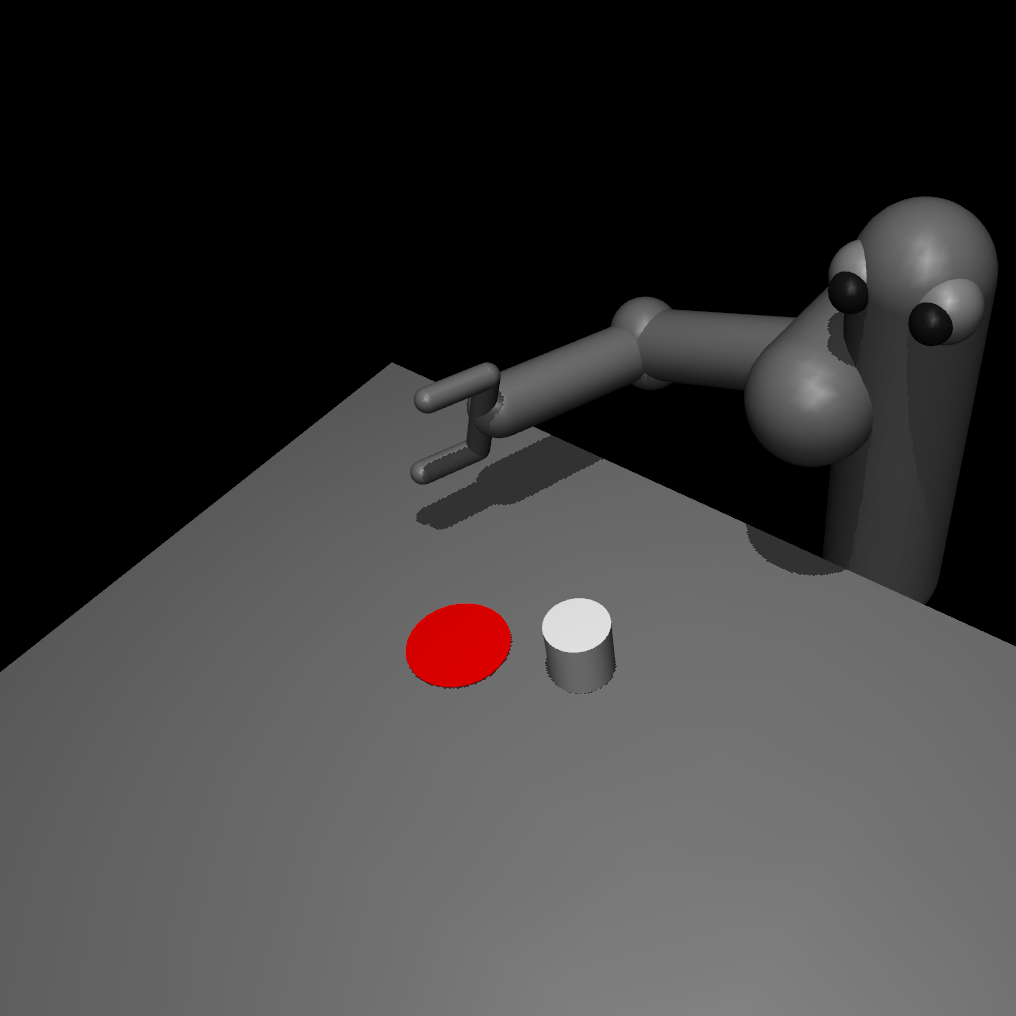}}
    \subfigure[Striker Task]{
        \includegraphics[clip,width=0.33\linewidth]{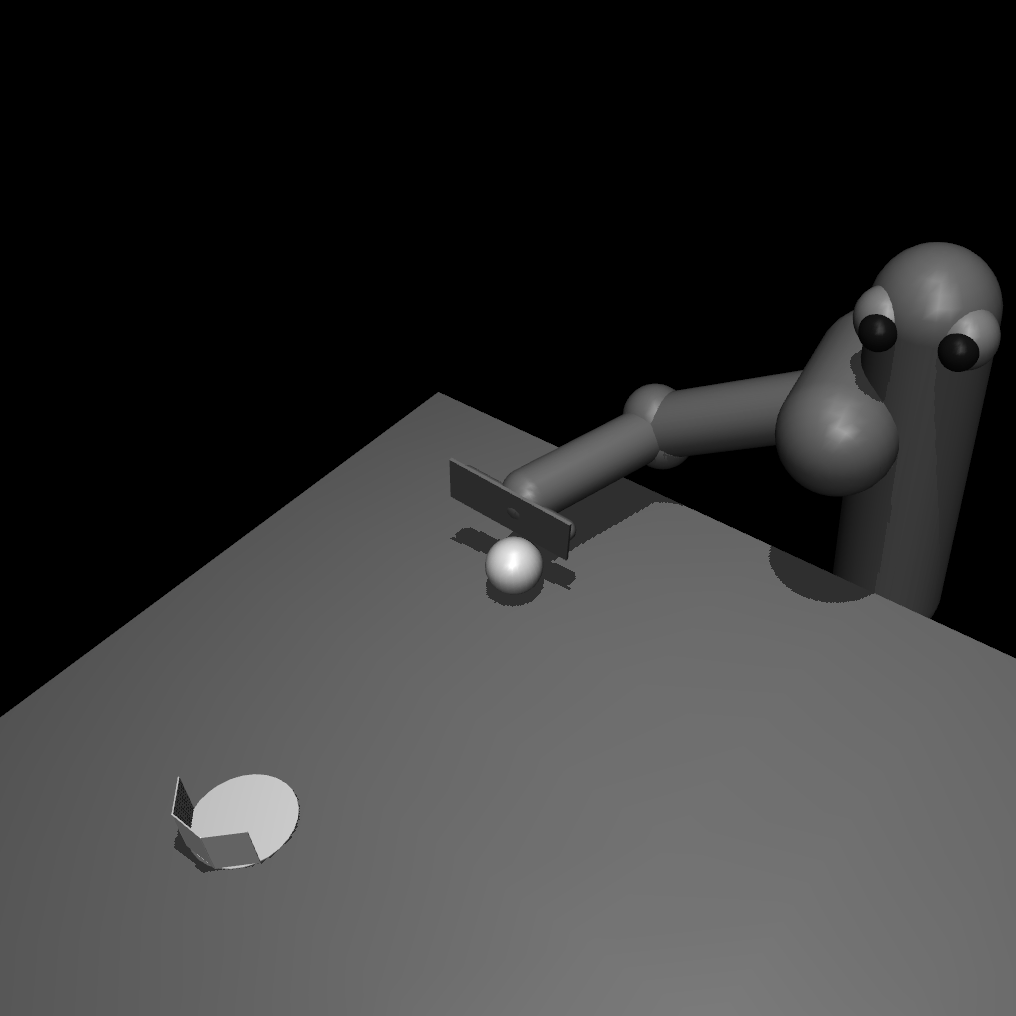}}
    \subfigure[Thrower Task]{
        \includegraphics[clip,width=0.33\linewidth]{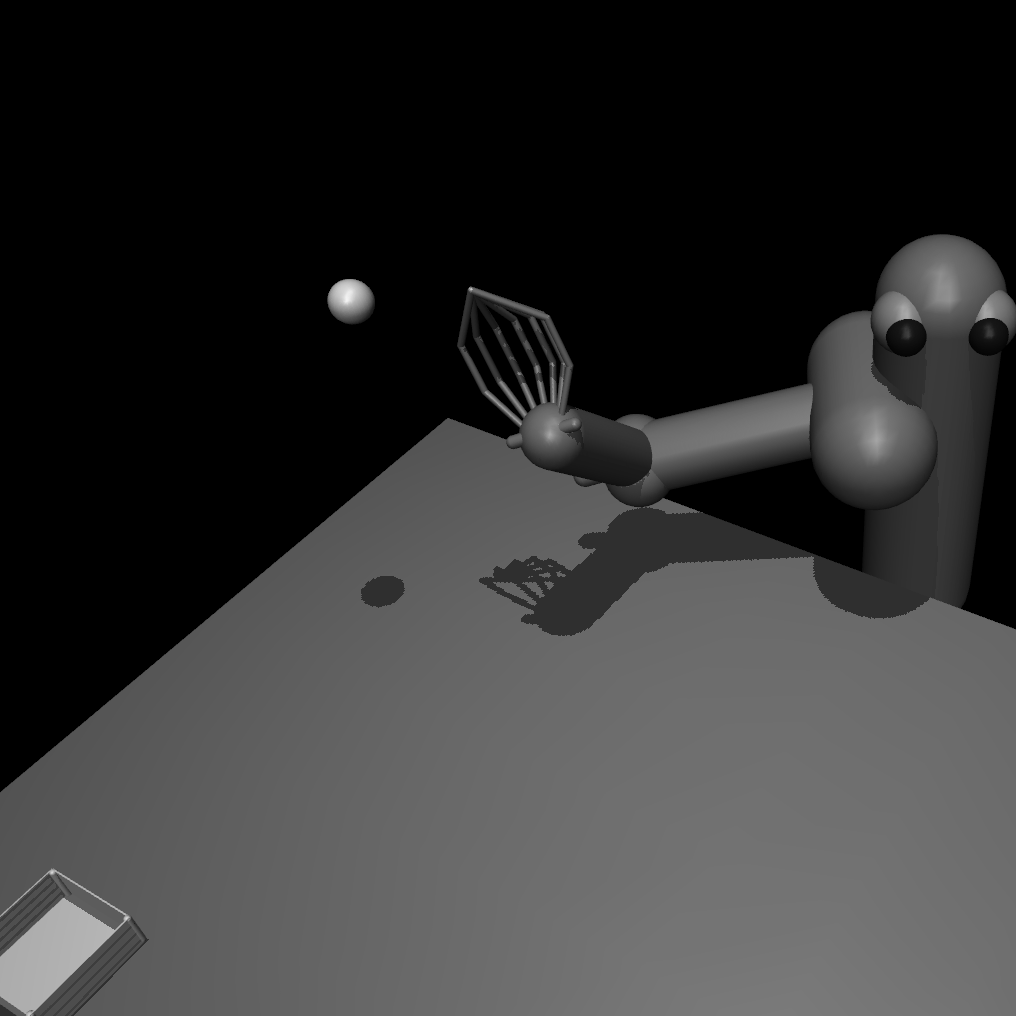}}
        \caption{
        \textbf{(a) Pusher }The purpose of the Pusher task is to control the robot arm to move a white cylindrical object into a red circle goal.
        In the initial state of the environment, the posture of the robot arm and the coordinates of the goal are fixed values, and the coordinates of the object are determined randomly.
        The task-achievement condition is that the XY coordinates of the object must fall within the goal range.
        
        \textbf{(b) Striker }The purpose of the Striker task is to control the robot arm and strike a white ball into the goal with a white fence.
        In the initial state of the environment, the posture of the robot arm and the coordinates of the object are fixed values, and the coordinates of the goal are determined randomly.
        The task-achievement condition is that the XY coordinates of the object must fall within the goal range.
        
        \textbf{(c) Thrower }The purpose of the Thrower task is to control the robot arm and throw a white ball into the white box goal.
        In the initial state of the environment, the posture of the robot arm and the coordinates of the object are fixed values, and the coordinates of the goal are determined randomly.
        The task-achievement condition is that the x and y coordinates of the object must be within the goal range, and the ball must be in contact with the bottom of the box.
        }
        \label{TaskImage}
\end{figure*}

We used three physics-based control tasks, which were performed by a 7-degree-of-freedom (DOF) manipulation robot.
Figure \ref{TaskImage} depicts each task.
In each task, the state space and action space are the same.
The action space is a 7-dimensional continuous value of the torque given to every seven joints of the arm.
The state space is a 23-dimensional continuous value, with the angles and angular velocities of the 7 joints, and the XYZ coordinates of objects, goals, and hands.

We generated expert behavior for these tasks by running PPO \cite{schulman2017proximal}. 
Each behavior is trained on these true reward functions defined in OpenAI Gym \cite{brockman2016openai} to generate expert policies.
Then, to evaluate the imitation performance, we sampled datasets of varying trajectory counts from expert policies.

We tested the proposed method against four baselines: PPO, BC, PPO+BC, and GAIL.
PPO+BC indicates that the PPO is learned using the policy pre-learned by BC as an initial parameter, which is the most standard approach to integrate IL and RL.
We trained the proposed method based on the task-achievement reward alone and did not use the reward defined in the OpenAI Gym \cite{brockman2016openai} to demonstrate that the only requirement for learning by our method is to define the task-achievement condition and collect expert trajectories.
This experiment aims to show that TRGAIL can learn a policy on only the task-achievement reward with the help of the demonstrators' trajectories.

\subsection{Results}

\begin{figure*}[!t]

    \includegraphics[clip,width=\linewidth]{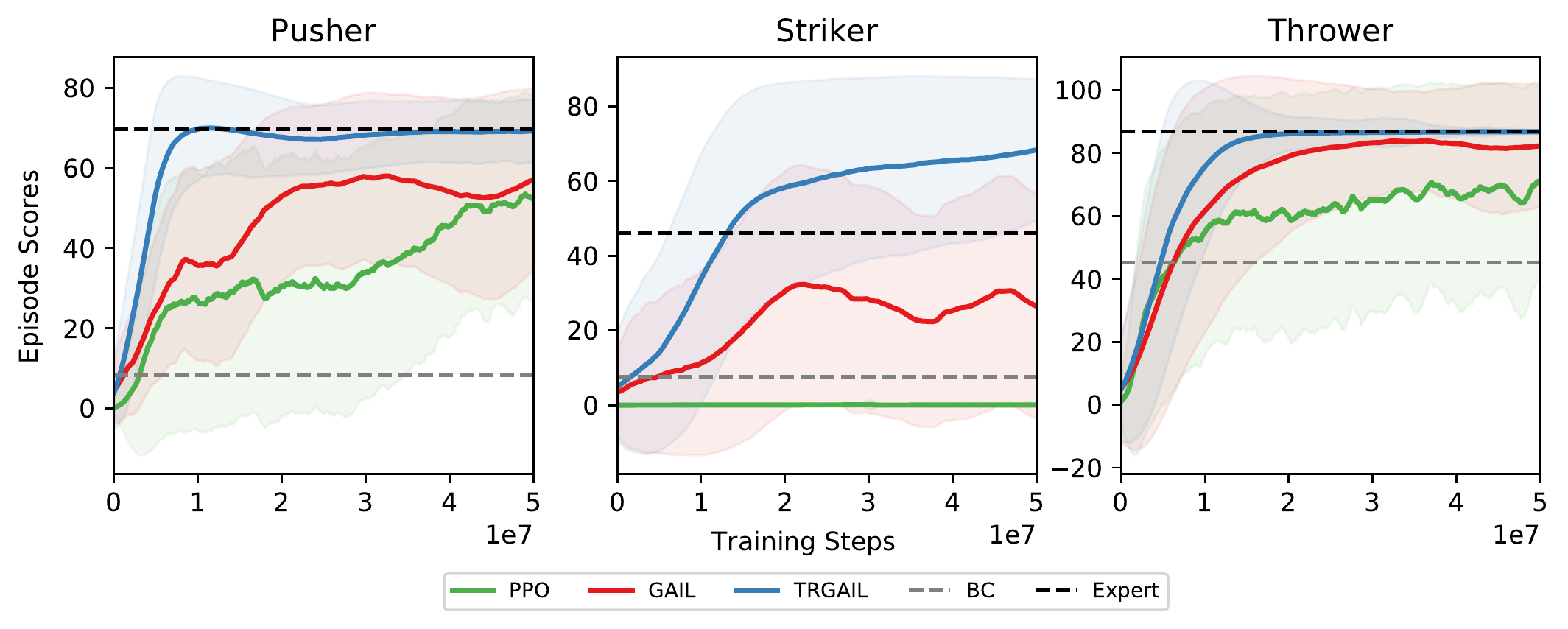}
    \caption{Performance of the learned policy considering that 15 expert trajectories are given to the agent in the Pusher Task, Striker Task, and Thrower Task.
    The x-axis represents the training time steps. One episode consists of 100 time steps. The y-axis represents the episode score, which is the number of time steps that achieved the goal in one episode.}
    \label{Score}
\end{figure*}

\begin{figure*}[!t]
        \includegraphics[clip,width=\linewidth]{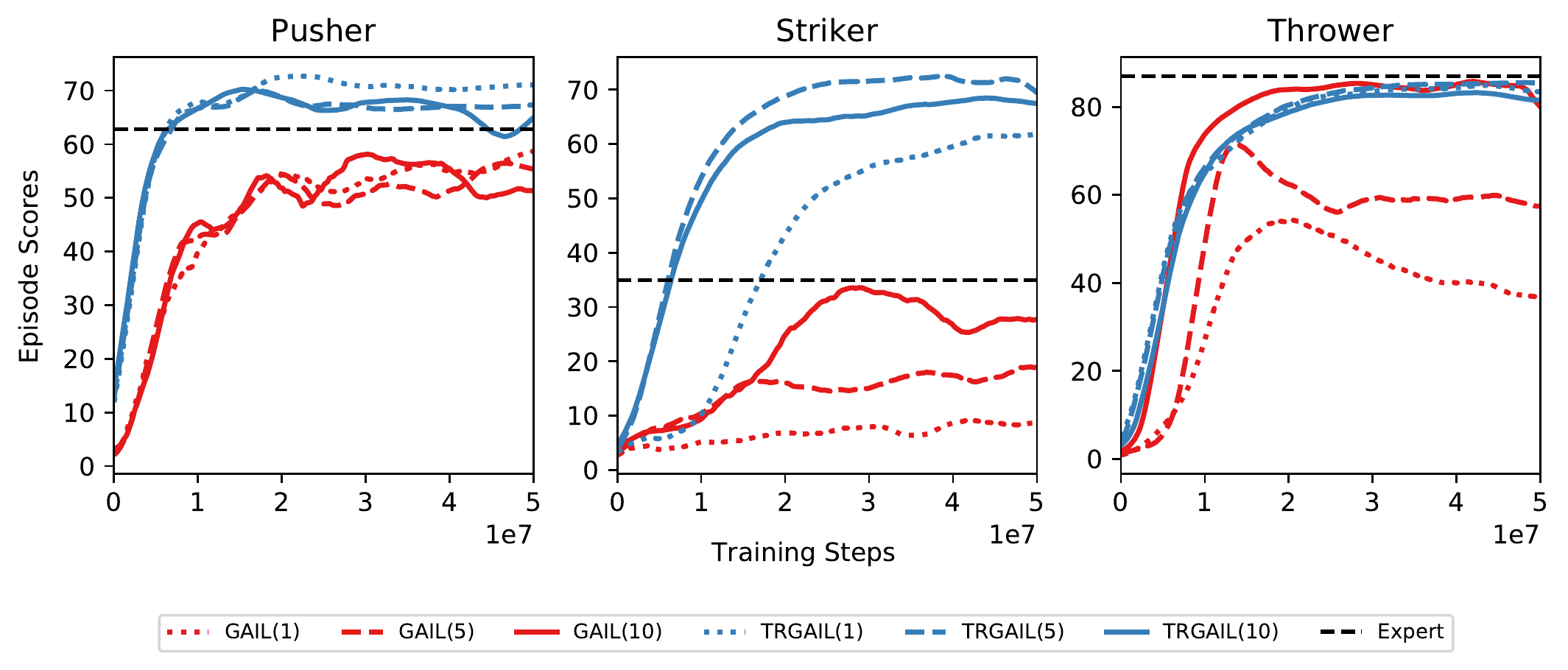}
        \caption{Performance of the learned policy in the case of changing the number of expert trajectories given to the agent in the Pusher Task, Striker Task, and Thrower Task.
The x-axis represents the training time steps. One episode consists of 100 time steps. The y-axis represents the episode score, which is the number of time steps that achieved the goal in one episode.}
        \label{Score_exp}
\end{figure*}

Figure \ref{Score} depicts the performance of the learned policies of each task in the MuJoCo physics simulator.
As shown in the figure, in most tasks, the proposed method can learn faster than the conventional method can, and the episode score of the learned policy is also higher.
In the Striker task, GAIL cannot efficiently learn the optimal policy because the expert trajectories trained by PPO are sub-optimal. In contrast, TRGAIL learned a better score than that of the sub-optimal trajectories because of the task-achievement reward.
In the Thrower task, considering that 15 expert trajectories were given, classical IL could sufficiently learn, and there was no significant difference between the proposed method and the conventional method.

Table \ref{Exp_result} and Figure \ref{Score_exp} present the experimental results when we change the number of expert trajectories given in each task.
We found that if the number of expert trajectories was decreased,  TRGAIL exhibited a higher learning performance than that exhibited by GAIL in most tasks, as GAIL could not learn enough.
\begin{table}[]
\begin{center}
\begin{tabular}{l|llll|llll|llll}
\hline
Task        & \multicolumn{4}{l|}{Pusher} & \multicolumn{4}{l|}{Striker} & \multicolumn{4}{l}{Thrower} \\ \hline
num of traj & 1     & 5     & 10   & 15   & 1     & 5     & 10    & 15   & 1     & 5    & 10    & 15   \\ \hline
Expert      & 72.0  & 70.0  & 69.7 & 69.0 & 43.0  & 41.2  & 46.2  & 38.7 & 87.0  & 87.0 & 87.0  & 87.0 \\
BC          & 0.0   & 1.4   & 8.4  & 34.0 & 0.0   & 2.5   & 7.6   & 2.3  & 0.1   & 26.0 & 45.2  & 63.5 \\
GAIL        & \underline{61.0}  & 58.4  & \underline{55.3} & 61.1 & 15.0  & 22.3  & 40.1  & 30.9 & 2.5  & 57.2 & 82.6  & \underline{86.1} \\
BC+PPO      & 49.2  & \underline{\bf74.6}  & 55.2  & \underline{69.3} & \underline{\bf{31.2}}  & \underline{56.5}  & \underline{56.1}  & \underline{56.7} & \underline{67.4}  & \underline{84.5} & \underline{84.9}  & 85.3     \\
TRGAIL      & \underline{\bf{65.7}}  & \underline{72.2}  & \underline{\bf{71.8}} & \underline{\bf72.4} & \underline{20.3}  & \underline{\bf{72.6}}  & \underline{\bf{72.4}}  & \underline{\bf67.8} & \underline{\bf{76.0}}  & \underline{\bf{86.6}} & \underline{\bf86.0}  & \underline{\bf86.9} \\ \hline
\end{tabular}

\caption{Experimental result of IL and combination approach of IL and RL in the case of changing the number of experts given in each task.
Episode scores indicate the number of time steps that achieved the goal in one episode.
The score value is the average of 100 episodes performed using the highest rated policy during the trial.}
\label{Exp_result}
\end{center}
\end{table}

\section{Conclusion}
In this study, we developed the new probabilistic graphical model framework for simultaneously observing multiple optimality emissions.
Furthermore, we demonstrated that the integrated learning method of RL and IL can be formulated as a probabilistic inference maximizing multiple optimalities to PGM with multiple optimality emission.
In experiments, our proposed method, which adapts generative adversarial imitation learning and task-achievement rewards to our framework, achieved significantly better performance than that achieved by the agents trained with RL or IL alone. 

However, we sometimes observed that the final performance of TRGAIL decreased as the number of experts given to the agent increased.
This is considered to be caused by the fact that IL inhibits the improvement of the score when a non-optimal expert is given.
Basically, RL has a low learning efficiency at the beginning of learning but can efficiently learn at a later stage where it can stably obtain a reward.
On the other hand, IL contributes significantly at the beginning of learning when no reward signal can be obtained, but at a later stage, it suffers a penalty for the distance from the expert.
From the above, we assume that it is preferable to learn while changing the weight parameter of RL and IL according to the progress of learning.
Specifically, approaches may be considered, such as setting a parameter of weight and decreasing this value as learning progresses. Our future study will focus on the formulation of this phenomenon in PGM.

Examining the relevance of other GAIL extension methods and our proposed method is also important.
The GAIL approach for obtaining desirable strategies by adding human-designed rewards has been previously proposed \cite{li2017infogail}.
Comparing these conventional approaches with our proposed method, the feature of the proposed method is to reformulate this approach as probabilistic inference on PGM.
By showing the relationship between such 
a problem and the probabilistic inference, various methods, such as existing probabilistic inference methods, can be applied.

Compared to the conventional IL, TRGAIL is superior in that it can be learned with a minimal number of experts.
Because it may be difficult to prepare many expert trajectories in IL, it is worth being able to learn with as few expert trajectories as possible.
In addition, because this method is applied to RL, it is an advantage that the expert trajectory to be given does not have to be optimal.
TRGAIL will learn well even when only an incomplete expert trajectory is given, which performs only part of the task.

In our future work, we will extend our proposed framework for hierarchical models.
In tasks for which the process to achieve the goal is complicated, the agent should segment a trajectory and learn the policy for each skill.
We need to propose hierarchical learning by extending the proposed framework using the connection with probabilistic inference proposed in this study.

Although this study focuses on the integration of RL and IL, the framework for simultaneously observing multiple optimality emissions is a general framework.
This framework could be applied in other tasks with the goal of maximizing with respect to multiple optimalities.

\label{lastpage}
\bibliographystyle{junsrt}
\bibliography{reference}

\begin{thebibliography}{10}

\bibitem{lin1992self}
Long-Ji Lin.
\newblock Self-improving reactive agents based on reinforcement learning,
  planning and teaching.
\newblock {\em Machine learning}, Vol.~8, No. 3-4, pp. 293--321, 1992.

\bibitem{lin1991programming}
Long~Ji Lin.
\newblock Programming robots using reinforcement learning and teaching.
\newblock In {\em Proceedings of the association for the advancement of
  artificial intelligence conference}, pp. 781--786, 1991.

\bibitem{silver2016mastering}
David Silver, Aja Huang, Chris~J Maddison, Arthur Guez, Laurent Sifre, George
  Van Den~Driessche, Julian Schrittwieser, Ioannis Antonoglou, Veda
  Panneershelvam, Marc Lanctot, et~al.
\newblock Mastering the game of go with deep neural networks and tree search.
\newblock {\em Nature}, Vol. 529, No. 7587, pp. 484--489, 2016.

\bibitem{brys2015reinforcement}
Tim Brys, Anna Harutyunyan, Halit~Bener Suay, Sonia Chernova, Matthew~E Taylor,
  and Ann Now{\'e}.
\newblock Reinforcement learning from demonstration through shaping.
\newblock In {\em Twenty-Fourth International Joint Conference on Artificial
  Intelligence}, 2015.

\bibitem{levine2013guided}
Sergey Levine and Vladlen Koltun.
\newblock Guided policy search.
\newblock In {\em International Conference on Machine Learning}, pp. 1--9,
  2013.

\bibitem{zhu2018reinforcement}
Yuke Zhu, Ziyu Wang, Josh Merel, Andrei Rusu, Tom Erez, Serkan Cabi, Saran
  Tunyasuvunakool, J{\'a}nos Kram{\'a}r, Raia Hadsell, Nando de~Freitas, et~al.
\newblock Reinforcement and imitation learning for diverse visuomotor skills.
\newblock {\em arXiv preprint arXiv:1802.09564}, 2018.

\bibitem{merel2017learning}
Josh Merel, Yuval Tassa, Sriram Srinivasan, Jay Lemmon, Ziyu Wang, Greg Wayne,
  and Nicolas Heess.
\newblock Learning human behaviors from motion capture by adversarial
  imitation.
\newblock {\em arXiv preprint arXiv:1707.02201}, 2017.

\bibitem{hamahata2008effective}
Keita Hamahata, Tadahiro Taniguchi, Kazutoshi Sakakibara, Ikuko Nishikawa,
  Kazuma Tabuchi, and Tetsuo Sawaragi.
\newblock Effective integration of imitation learning and reinforcement
  learning by generating internal reward.
\newblock In {\em 2008 Eighth International Conference on Intelligent Systems
  Design and Applications}, Vol.~3, pp. 121--126, 2008.

\bibitem{le2018hierarchical}
Nan Jiang Alekh Agarwal Miroslav Dud^^c3^^adk Yisong~Yue Le, Hoang~Minh and Hal
  Daum^^c3^^a9.
\newblock Hierarchical imitation and reinforcement learning.
\newblock {\em International Conference on Machine Learning}, 2018.

\bibitem{nair2018overcoming}
Ashvin Nair, Bob McGrew, Marcin Andrychowicz, Wojciech Zaremba, and Pieter
  Abbeel.
\newblock Overcoming exploration in reinforcement learning with demonstrations.
\newblock In {\em 2018 IEEE International Conference on Robotics and Automation
  (ICRA)}, pp. 6292--6299, 2018.

\bibitem{hester2018deep}
Todd Hester, Matej Vecerik, Olivier Pietquin, Marc Lanctot, Tom Schaul, Bilal
  Piot, Dan Horgan, John Quan, Andrew Sendonaris, Ian Osband, et~al.
\newblock Deep q-learning from demonstrations.
\newblock In {\em Thirty-Second AAAI Conference on Artificial Intelligence},
  2018.

\bibitem{andreas2017modular}
Jacob Andreas, Dan Klein, and Sergey Levine.
\newblock Modular multitask reinforcement learning with policy sketches.
\newblock In {\em Proceedings of the 34th International Conference on Machine
  Learning-Volume 70}, pp. 166--175, 2017.

\bibitem{sun2017deeply}
Wen Sun, Arun Venkatraman, Geoffrey~J Gordon, Byron Boots, and J~Andrew
  Bagnell.
\newblock Deeply aggrevated: Differentiable imitation learning for sequential
  prediction.
\newblock In {\em Proceedings of the 34th International Conference on Machine
  Learning-Volume 70}, pp. 3309--3318, 2017.

\bibitem{rhinehart2018deep}
Nicholas Rhinehart, Rowan McAllister, and Sergey Levine.
\newblock Deep imitative models for flexible inference, planning, and control.
\newblock {\em arXiv preprint arXiv:1810.06544}, 2018.

\bibitem{attias2003planning}
Hagai Attias.
\newblock Planning by probabilistic inference.
\newblock In {\em Proceedings of the 9th International Workshop on Artificial
  Intelligence and Statistics.}, 2003.

\bibitem{todorov2007linearly}
Emanuel Todorov.
\newblock Linearly-solvable markov decision problems.
\newblock In {\em Advances in neural information processing systems}, pp.
  1369--1376, 2007.

\bibitem{kappen2012optimal}
Hilbert~J Kappen, Vicen{\c{c}} G{\'o}mez, and Manfred Opper.
\newblock Optimal control as a graphical model inference problem.
\newblock {\em Machine learning}, Vol.~87, No.~2, pp. 159--182, 2012.

\bibitem{toussaint2009robot}
Marc Toussaint.
\newblock Robot trajectory optimization using approximate inference.
\newblock In {\em Proceedings of the 26th annual international conference on
  machine learning}, pp. 1049--1056, 2009.

\bibitem{rawlik2013stochastic}
Konrad Rawlik, Marc Toussaint, and Sethu Vijayakumar.
\newblock On stochastic optimal control and reinforcement learning by
  approximate inference.
\newblock In {\em Twenty-Third International Joint Conference on Artificial
  Intelligence}, 2013.

\bibitem{ziebart2010modeling}
Brian~D Ziebart, J~Andrew Bagnell, and Anind~K Dey.
\newblock Modeling interaction via the principle of maximum causal entropy.
\newblock 2010.

\bibitem{ho2016generative}
Jonathan Ho and Stefano Ermon.
\newblock Generative adversarial imitation learning.
\newblock In {\em Advances in Neural Information Processing Systems}, pp.
  4565--4573, 2016.

\bibitem{levine2018reinforcement}
Sergey Levine.
\newblock Reinforcement learning and control as probabilistic inference:
  Tutorial and review.
\newblock {\em arXiv preprint arXiv:1805.00909}, 2018.

\bibitem{mnih2013playing}
Volodymyr Mnih, Koray Kavukcuoglu, David Silver, Alex Graves, Ioannis
  Antonoglou, Daan Wierstra, and Martin Riedmiller.
\newblock Playing atari with deep reinforcement learning.
\newblock {\em In Deep Learning, Neural Information Processing Systems
  Workshop}, 2013.

\bibitem{mnih2015human}
Volodymyr Mnih, Koray Kavukcuoglu, David Silver, Andrei~A Rusu, Joel Veness,
  Marc~G Bellemare, Alex Graves, Martin Riedmiller, Andreas~K Fidjeland, Georg
  Ostrovski, et~al.
\newblock Human-level control through deep reinforcement learning.
\newblock {\em Nature}, Vol. 518, No. 7540, pp. 529--533, 2015.

\bibitem{abdolmaleki2018maximum}
Abbas Abdolmaleki, Jost~Tobias Springenberg, Yuval Tassa, Remi Munos, Nicolas
  Heess, and Martin Riedmiller.
\newblock Maximum a posteriori policy optimisation.
\newblock {\em arXiv preprint arXiv:1806.06920}, 2018.

\bibitem{fu2018variational}
Justin Fu, Avi Singh, Dibya Ghosh, Larry Yang, and Sergey Levine.
\newblock Variational inverse control with events: A general framework for
  data-driven reward definition.
\newblock In {\em Advances in Neural Information Processing Systems}, pp.
  8538--8547, 2018.

\bibitem{bain1999framework}
Michael Bain and Claude Sommut.
\newblock A framework for behavioural claning.
\newblock {\em Machine intelligence}, Vol.~15, No.~15, p. 103, 1999.

\bibitem{pomerleau1991efficient}
Dean~A Pomerleau.
\newblock Efficient training of artificial neural networks for autonomous
  navigation.
\newblock {\em Neural Computation}, Vol.~3, No.~1, pp. 88--97, 1991.

\bibitem{bojarski2016end}
Mariusz Bojarski, Davide Del~Testa, Daniel Dworakowski, Bernhard Firner, Beat
  Flepp, Prasoon Goyal, Lawrence~D Jackel, Mathew Monfort, Urs Muller, Jiakai
  Zhang, et~al.
\newblock End to end learning for self-driving cars.
\newblock {\em arXiv preprint arXiv:1604.07316}, 2016.

\bibitem{nakanishi2004learning}
Jun Nakanishi, Jun Morimoto, Gen Endo, Gordon Cheng, Stefan Schaal, and Mitsuo
  Kawato.
\newblock Learning from demonstration and adaptation of biped locomotion.
\newblock {\em Robotics and autonomous systems}, Vol.~47, No. 2-3, pp. 79--91,
  2004.

\bibitem{kalakrishnan2009learning}
Mrinal Kalakrishnan, Jonas Buchli, Peter Pastor, and Stefan Schaal.
\newblock Learning locomotion over rough terrain using terrain templates.
\newblock In {\em Intelligent Robots and Systems, 2009. IROS 2009. IEEE/RSJ
  International Conference on}, pp. 167--172, 2009.

\bibitem{ziebart2008maximum}
Brian~D Ziebart, Andrew~L Maas, J~Andrew Bagnell, and Anind~K Dey.
\newblock Maximum entropy inverse reinforcement learning.
\newblock In {\em Proceedings of the association for the advancement of
  artificial intelligence conference}, Vol.~8, pp. 1433--1438, 2008.

\bibitem{wulfmeier2015maximum}
Markus Wulfmeier, Peter Ondruska, and Ingmar Posner.
\newblock Maximum entropy deep inverse reinforcement learning.
\newblock {\em arXiv preprint arXiv:1507.04888}, 2015.

\bibitem{finn2016guided}
Chelsea Finn, Sergey Levine, and Pieter Abbeel.
\newblock Guided cost learning: Deep inverse optimal control via policy
  optimization.
\newblock In {\em International Conference on Machine Learning}, pp. 49--58,
  2016.

\bibitem{fu2017learning}
Justin Fu, Katie Luo, and Sergey Levine.
\newblock Learning robust rewards with adversarial inverse reinforcement
  learning.
\newblock {\em InInternational Conference on Learning Representations (ICLR)},
  2018.

\bibitem{goodfellow2014generative}
Ian Goodfellow, Jean Pouget-Abadie, Mehdi Mirza, Bing Xu, David Warde-Farley,
  Sherjil Ozair, Aaron Courville, and Yoshua Bengio.
\newblock Generative adversarial nets.
\newblock In {\em Advances in neural information processing systems}, pp.
  2672--2680, 2014.

\bibitem{schulman2015trust}
John Schulman, Sergey Levine, Pieter Abbeel, Michael Jordan, and Philipp
  Moritz.
\newblock Trust region policy optimization.
\newblock In {\em Proceedings of the 32nd International Conference on Machine
  Learning (ICML-15)}, pp. 1889--1897, 2015.

\bibitem{schulman2017proximal}
John Schulman, Filip Wolski, Prafulla Dhariwal, Alec Radford, and Oleg Klimov.
\newblock Proximal policy optimization algorithms.
\newblock {\em arXiv preprint arXiv:1707.06347}, 2017.

\bibitem{kingma2014adam}
Diederik~P Kingma and Jimmy Ba.
\newblock Adam: A method for stochastic optimization.
\newblock {\em arXiv preprint arXiv:1412.6980}, 2014.

\bibitem{baram2017end}
Nir Baram, Oron Anschel, Itai Caspi, and Shie Mannor.
\newblock End-to-end differentiable adversarial imitation learning.
\newblock In {\em Proceedings of the 34th International Conference on Machine
  Learning-Volume 70}, pp. 390--399, 2017.

\bibitem{li2017infogail}
Yunzhu Li, Jiaming Song, and Stefano Ermon.
\newblock Infogail: Interpretable imitation learning from visual
  demonstrations.
\newblock In {\em Advances in Neural Information Processing Systems}, pp.
  3815--3825, 2017.

\bibitem{hausman2017multi}
Karol Hausman, Yevgen Chebotar, Stefan Schaal, Gaurav Sukhatme, and Joseph~J
  Lim.
\newblock Multi-modal imitation learning from unstructured demonstrations using
  generative adversarial nets.
\newblock In {\em Advances in Neural Information Processing Systems}, pp.
  1235--1245, 2017.

\bibitem{lin2018acgail}
Jiahao Lin and Zongzhang Zhang.
\newblock Acgail: Imitation learning about multiple intentions with auxiliary
  classifier gans.
\newblock In {\em Pacific Rim International Conference on Artificial
  Intelligence}, pp. 321--334, 2018.

\bibitem{todorov2012mujoco}
Emanuel Todorov, Tom Erez, and Yuval Tassa.
\newblock Mujoco: A physics engine for model-based control.
\newblock In {\em Intelligent Robots and Systems (IROS), 2012 IEEE/RSJ
  International Conference on}, pp. 5026--5033, 2012.

\bibitem{brockman2016openai}
Greg Brockman, Vicki Cheung, Ludwig Pettersson, Jonas Schneider, John Schulman,
  Jie Tang, and Wojciech Zaremba.
\newblock Openai gym.
\newblock {\em arXiv preprint arXiv:1606.01540}, 2016.

\end{thebibliography}
\end{document}


\appendix
\section{Experiment Details}
\begin{table}[h]
    \centering
    \caption{Pusher環境の行動空間}
    \begin{minipage}[t]{.7\textwidth}
        \begin{tabular}{lll}
            index & name & range \\
            \hline
            1 & r\_shoulder\_pan\_joint     & {[}-2,2{]} \\
            2 & r\_shoulder\_lift\_joint    & {[}-2,2{]} \\
            3 & r\_upper\_arm\_roll\_joint  & {[}-2,2{]} \\
            4 & r\_elbow\_flex\_joint       & {[}-2,2{]} \\
            5 & r\_forearm\_roll\_joint     & {[}-2,2{]} \\
            6 & r\_wrist\_flex\_joint       & {[}-2,2{]} \\
            7 & r\_wrist\_roll\_joint       & {[}-2,2{]} \\ \hline
        \end{tabular}
    \end{minipage}
    \begin{minipage}[t]{.4\textwidth}
        \begin{tabular}{lll}
            index & name & range \\
            \hline
            1-7   & 7関節の角度                     &                        \\
                  & r\_shoulder\_pan\_joint & {[}-2.2854,1.714602{]} \\
                  & r\_shoulder\_lift\_joint& {[}-0.5236,1.3963{]}   \\
                  & r\_upper\_arm\_roll\_joint& {[}-1.5,1.7{]}         \\
                  & r\_elbow\_flex\_joint   & {[}-2.3213,0{]}        \\
                  & r\_forearm\_roll\_joint & {[}-1.5,1.5{]}         \\
                  & r\_wrist\_flex\_joint   & {[}-1.094,0{]}         \\
                  & r\_wrist\_roll\_joint   & {[}-1.5,1.5{]}         \\
            8-14  & 7関節の角速度                    &                        \\
            15-17 & 手先のx,y,z座標                 &                        \\
            18-20 & オブジェクトのx,y,z座標             &                        \\
            21-23 & ゴールのx,y,z座標                &                        \\ \hline
        \end{tabular}
    \end{minipage}
\end{table}